\documentclass[10pt,twocolumn,letterpaper]{article}

\newcommand{\ours}{\textsc{SurvPath}}
\newcommand{\pathways}{\mathcal{P}}
\newcommand{\dataset}{\mathcal{D}}
\newcommand{\genes}{\mathcal{G}}
\newcommand{\histology}{\mathcal{H}}
\newcommand{\real}{\mathbb{R}}

\usepackage{cvpr}              

\usepackage{url}
\usepackage{cite}
\usepackage{amsmath,amssymb,amsfonts}
\usepackage{algorithmic}
\usepackage{graphicx}
\usepackage{textcomp}
\usepackage{paralist}
\usepackage{listings}
\usepackage{booktabs}
\usepackage{hhline}
\usepackage{framed,multirow}
\usepackage{lipsum}  
\usepackage{soul}
\usepackage{cancel}
\usepackage{booktabs}
\usepackage{latexsym}
\usepackage{array}
\usepackage{stackengine}
\usepackage{relsize}
\usepackage[ruled,vlined]{algorithm2e}

\makeatletter
\renewcommand*{\@fnsymbol}[1]{%
  \ensuremath{%
    \ifcase#1\or *
    \or \dagger
    \or \ddagger 
    \or \mathsection
    \or \mathparagraph
    \or \|\or **\or \dagger\dagger
    \or \ddagger\ddagger \else\@ctrerr\fi}%
}
\makeatother

\usepackage[dvipsnames]{xcolor}

\definecolor{cvprblue}{rgb}{0.21,0.49,0.74}
\usepackage[pagebackref,breaklinks,colorlinks,citecolor=cvprblue]{hyperref}

\def\paperID{10446} 
\def\confName{CVPR}
\def\confYear{2024}

\title{Modeling Dense Multimodal Interactions Between Biological Pathways and Histology for Survival Prediction}

\author{Guillaume Jaume$^{1,2,}$\thanks{Equal contribution}, Anurag Vaidya$^{1,2,*}$, Richard J. Chen$^{1,2}$, Drew F.K. Williamson$^{1,2,}$\thanks{Presently at Emory University School of Medicine} \\
Paul Pu Liang$^{3}$, Faisal Mahmood$^{1,2}$\\
${^1}$Mass General Brigham, ${^2}$Harvard University,  ${^3}$CMU\\
{\tt\small gjaume@bwh.harvard.edu, ajvaidya@bwh.harvard.edu, faisalmahmood@bwh.harvard.edu}
}

\begin{document}

\maketitle
\begin{abstract}
Integrating whole-slide images (WSIs) and bulk transcriptomics for predicting patient survival can improve our understanding of patient prognosis. However, this multimodal task is particularly challenging due to the different nature of these data: WSIs represent a very high-dimensional spatial description of a tumor, while bulk transcriptomics represent a global description of gene expression levels within that tumor. In this context, our work aims to address two key challenges: (1) how can we tokenize transcriptomics in a semantically meaningful and interpretable way?, and (2) how can we capture dense multimodal interactions between these two modalities? Here, we propose to learn biological pathway tokens from transcriptomics that can encode specific cellular functions. Together with histology patch tokens that encode the slide morphology, we argue that they form appropriate reasoning units for interpretability. We fuse both modalities using a memory-efficient multimodal Transformer that can model interactions between pathway and histology patch tokens. Our model, $\ours$, achieves state-of-the-art performance when evaluated against unimodal and multimodal baselines on five datasets from The Cancer Genome Atlas. Our interpretability framework identifies key multimodal prognostic factors, and, as such, can provide valuable insights into the interaction between genotype and phenotype. Code available at \href{https://github.com/mahmoodlab/SurvPath}{https://github.com/mahmoodlab/SurvPath}. 
\end{abstract}

\section{Introduction}
\label{sec:intro}
 
\begin{figure}[t]
    \centering
    \includegraphics[width=0.5\textwidth]{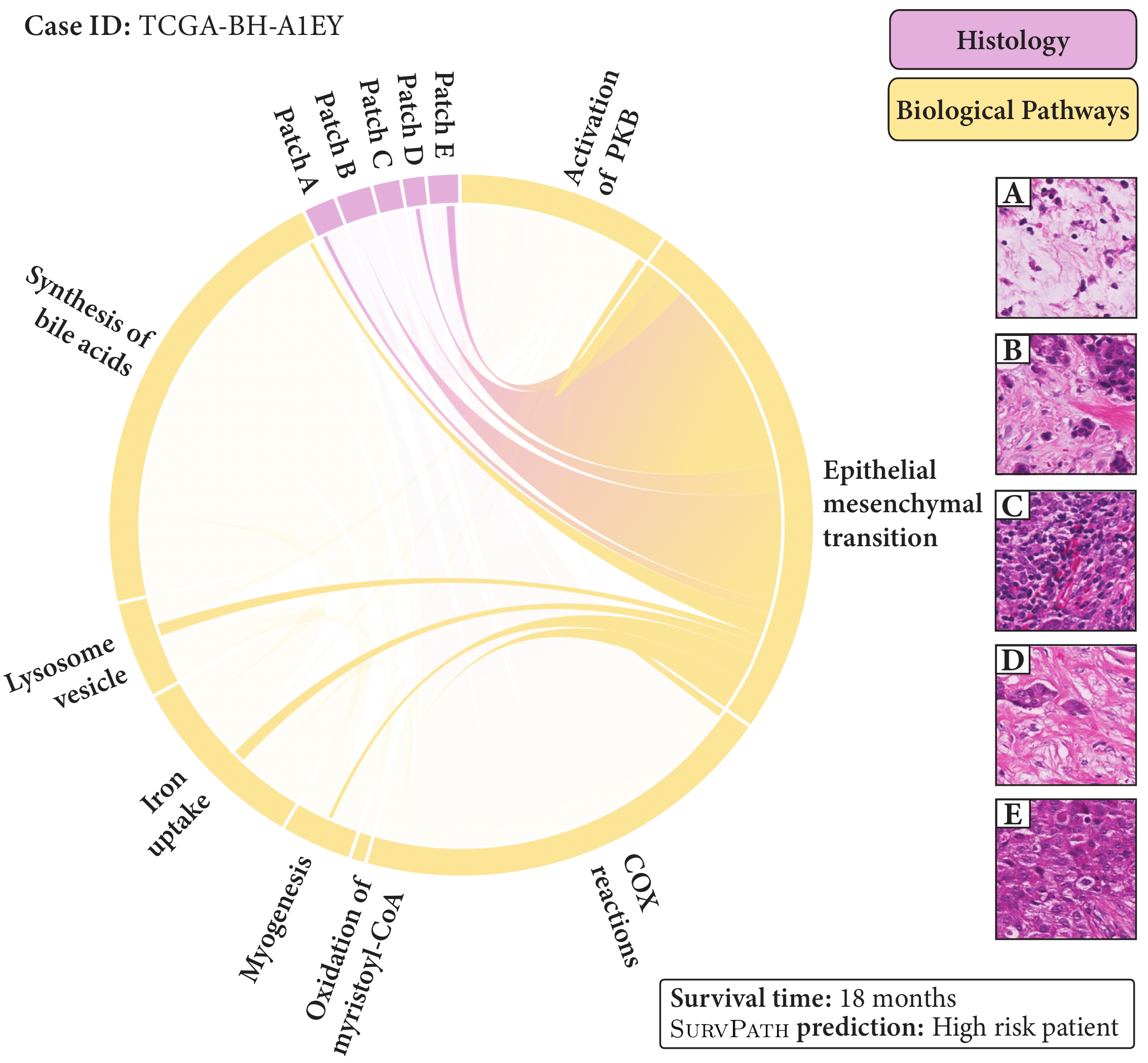}
    \caption{\textbf{Multimodal interpretability with $\ours$.} $\ours$ enables visualization of multimodal interactions via a Transformer cross-attention between \emph{biological pathways} and \emph{morphological patterns}, here exemplified in a high-risk breast cancer. The chord thickness denotes attention weight.
    }
    \label{fig:catchy}
\end{figure}

Predicting patient prognosis is a fundamental task in computational pathology (CPATH) that aims to utilize histology whole-slide images (WSIs) for automated risk assessment, patient stratification, and response-to-treatment prediction~\cite{song2023artificial, coudray2018classification,bandi2018detection,campanella2019clinical,bulten2022artificial,shmatko2022artificial}. Patient prognostication is often framed as a survival task, in which the goal is to learn risk estimates that correctly rank the survival time from the primary diagnostic WSI(s) ~\cite{yao2020whole,wulczyn2018deep,kather2019predicting,wang2021predicting,lee2022derivation}. As WSIs can be as large as 100,000 $\times$ 100,000 pixels, weakly supervised methods such as multiple instance learning (MIL) are often employed for survival prediction. In MIL, WSIs are tokenized into small patches, from which features are extracted and fed into pooling networks, such as attention networks, for downstream classification~\cite{ilse2018attention,shao2021transmil}.

While histology provides phenotypic information about cell types and their organization into tissues, alternate modalities can provide complementary signals that may independently be linked to prognosis. For instance, bulk transcriptomics, which represents the average gene expression in a tissue, can reveal a richer global landscape of cell types and cell states~\cite{li2022hfbsurv,wang2021GPDBN} and has been shown to be a strong predictor of patient survival~\cite{gyorffy2021survival,nagy2020pancancer,raghavan2021microenvironment}. By combining both modalities, we can integrate the global information provided by bulk transcriptomics with the spatial information from the WSI. While most existing methods adopt \emph{late fusion} mechanisms~\cite{chen2022pan,li2022hfbsurv} (\emph{i.e.,} fusing modality-level representations), we design an \emph{early fusion} method that can explicitly model fine-grained cross-modal relationships between local morphological patterns and transcriptomics. In comparison with widely employed vision-language models~\cite{radford2021learning,alayrac2022flamingo, wang2022image}, multimodal fusion of transcriptomics and histology presents two key technical challenges:

1. \emph{Tokenizing transcriptomics modality:} Modalities based on image and text can be unequivocally tokenized into object regions and word tokens~\cite{liang2022foundations,wang2022image}, however, tokenizing transcriptomics in a semantically meaningful and interpretable way is challenging. As transcriptomics data is already naturally represented as a feature vector, many prior studies ignore tokenization and directly concatenate the entire feature with other modalities, which limits multimodal learning to \emph{late fusion} operations~\cite{wang2021GPDBN,li2022hfbsurv}. Alternatively, genes can be partitioned into coarse functional sets that represent different gene families (\emph{e.g.,} tumor-suppressor genes and oncogenes) that can be used as tokens~\cite{chen2021multimodal}. Nevertheless, such sets provide a rudimentary and incomplete depiction of intracellular interactions as one gene family can be involved in different cellular functions. Consequently, they may lack semantic correspondence with fine-grained morphologies. Instead, we propose tokenizing genes according to established \emph{biological pathways}~\cite{subramanian2005gene,liberzon2015the,gillespie2021reactome}. Pathways are gene sets with known interactions that relate to \emph{specific} cellular functions, such as the TGF-$\beta$ signaling cascade, which contributes to the epithelial-mesenchymal transition in breast cancer~\cite{wang2013epithelial}. Compared to coarse sets (\emph{e.g.,} $N_{\pathways} = 6$~\cite{chen2021multimodal}), pathway-based gene grouping can yield hundreds to thousands of tokens that represent unique molecular processes ($N_{\pathways} = 331$ in our work), which we hypothesize are more suitable representations for multimodal fusion with histology. In addition, as pathways represent unique cellular functions, they constitute appropriate basic reasoning units for interpretability (see Fig.~\ref{fig:catchy}).

2. \emph{Capturing dense multimodal interactions:} Early fusion of histology and pathway tokens can be done with a  Transformer that uses self-attention to capture pairwise similarities between all tokens~\cite{vaswani2017attention}. However, modeling pairwise interactions between large sets of histology patch tokens (\emph{e.g.,} $N_{\histology} = 15,000$) and pathway tokens ($N_{\pathways} = 331$) poses scalability challenges for fusion. Due to the quadratic complexity of the Transformer attention, modeling all possible interactions imposes substantial computational and memory requirements. To tackle this issue, we introduce a new unified, memory-efficient attention mechanism that can model patch-to-pathway, pathway-to-patch, and pathway-to-pathway interactions. Modeling these three forms of interaction is achieved by the following: (1) designing the queries, keys, and values to share parameters across token types~\cite{jaegle2021perceiver,jaegle2022perceiver}, and (2) simplifying the attention layer to ignore patch-to-patch interactions, which we find through experimentation to be not as effective for survival analysis.

To summarize, our contributions are (1) a transcriptomics tokenizer that leverages existing knowledge of cellular biology to generate \emph{biological pathway} tokens; (2) $\ours$, a memory-efficient and resolution agnostic, multimodal Transformer formulation that integrates transcriptomics and patch tokens for predicting patient survival; (3) a multi-level interpretability framework that enables deriving unimodal and cross-modal insights about the prediction; (4) a series of experiments and ablations showing the predictive power of $\ours$, using five datasets from The Cancer Genome Atlas Program (TCGA) and benchmarked against both unimodal and multimodal fusion methods.

\begin{figure*}[t]
    \centering
    \includegraphics[width=0.95\textwidth]{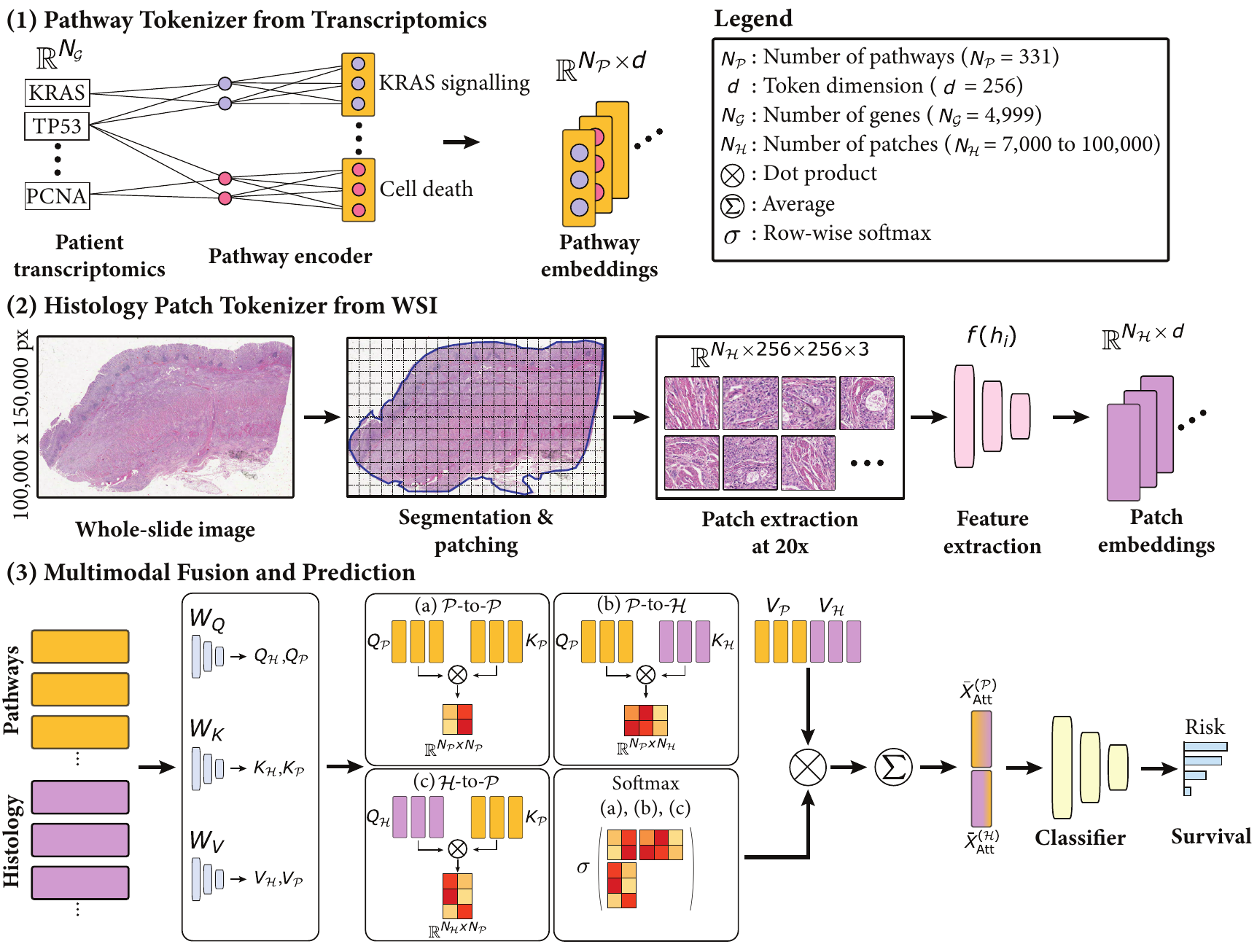}
    \caption{\textbf{Block diagram of $\ours$.} (1) We tokenize transcriptomics into \emph{biological pathway} tokens that are semantically meaningful, interpretable, and end-to-end learnable. (2) We further tokenize the corresponding histology whole-slide image into patch tokens using an SSL pre-trained feature extractor. (3) We combine pathway and patch tokens using a memory-efficient multimodal Transformer for survival outcome prediction.}
    \label{fig:block_diagram}
\end{figure*}

\section{Related Work}

\subsection{Survival Analysis on WSIs}

Recently, several histology-based survival models have been proposed~\cite{zhu2017wsisa,yao2020whole,wetstein2022deep,shao2023hvtsurv,song2024morphological}. Most contributions have been dedicated to modeling tumor heterogeneity and the tumor microenvironment using MIL. To this end, several MIL pooling strategies have been proposed, such as using graph neural networks to model local patch interactions~\cite{mackenzie2022neural,di2022generating,lee2022derivation}, accounting for the variance between patch embeddings~\cite{schirris2022deepsmile}, or adopting multi-magnification patch representations~\cite{liu2022deep}.

\subsection{Multimodal Transformers and Interpretability}

In parallel, the use of Transformers for multimodal fusion has gained significant attention in classification and generative tasks~\cite{tsai2019multimodal,xu2022multimodal,shamshad2022Transformers}. Multimodal tokens can be concatenated and fed to a regular Transformer~\cite{vaswani2017attention,dosovitskiy2020image}, a hierarchical Transformer~\cite{lin2020interbert}, or a cross-attention Transformer~\cite{nguyen2018improved,lu2019vilbert,murahari2020large}. As the number and dimensionality of modalities increase, the typical sequence length can become too large to be fed to vanilla Transformers, hence the need for low-complexity methods. Several models have proposed re-formulations of self-attention to reduce memory and computational requirements~\cite{beltagy2020longformer,xiong2021nystrom,jaegle2021perceiver,choromanski2020rethinking,wu2022flowformer,hua22Transformer,dao2022flashattention,dao2023flashattention2}, for instance, by approximating self-attention with a low-rank decomposition~\cite{lu2021soft,xiong2021nystrom}, using latent bottleneck distillation~\cite{jaegle2021perceiver,nagrani2021attention,jaegle2022perceiver}, by optimizing GPU reads/writes~\cite{dao2022flashattention,dao2023flashattention2} or using sparse attention patterns~\cite{beltagy2020longformer,qiu2020blockwise}. Recently, interpretable multimodal models or post-hoc interpretation methods~\cite{liang2022multiviz,wang2021m2lens,aflalo2022vl} have also emerged as a critical area of research, especially in sensitive human-AI collaborative decision-making scenarios such as healthcare and human-computer interactions.

\subsection{Multimodal Survival Analysis}

Multimodal integration is an important objective in cancer prognosis~\cite{shmatko2022artificial}, as combining histology and omics data such as genomics or transcriptomics is the current clinical practice for many cancer types. The majority of these works employ \emph{late fusion} mechanisms~\cite{valesilva2020multisurv,chen2020pathomic}, and mostly differ in the way modality fusion is operated. Fusion can be based on vector concatenation~\cite{mobadersany2018predicting}, modality-level alignment~\cite{cheerla2019deep}, bilinear pooling (\emph{i.e.,} Kronecker product)~\cite{chen2020pathomic,wang2021GPDBN}, or factorized bilinear pooling~\cite{li2022hfbsurv,qiu2023deep}. 

Differently, \emph{early fusion} mechanisms can be employed, in which cross-modal interactions between individual constituents of the input are modeled~\cite{chen2021multimodal,zhou2023cross,ding2023pathology,xu2023multimodal}. Our work builds off MCAT~\cite{chen2021multimodal}, which uses a cross-attention module to model the attention of histology patches (keys, values) toward gene sets (queries). However, MCAT has several limitations: (1) cross-attention being one-sided and models only patch-to-genes interactions, (2) transcriptomics tokenization using coarse sets that do not reflect actual molecular processes, and (3) significant gene overlap between sets, which leads to redundant cross-attention heatmaps.


\section{Method}

Here, we present $\ours$, our proposed method for multimodal survival prediction based on histology and transcriptomics. Sec.~\ref{sec:omics_tokenizer} presents the transcriptomics encoder to build biological pathway tokens, Sec.~\ref{sec:histology_tokenizer} presents the histology encoder to build patch tokens, Sec.~\ref{sec:multimodal} presents our Transformer-based multimodal aggregation, and Sec.~\ref{sec:survival} presents its application to survival prediction (see Fig.~\ref{fig:block_diagram}). Finally, Sec.~\ref{sec:multilevel_inter} introduces our multi-level interpretability framework.

\subsection{Pathway Tokenizer from Transcriptomics} \label{sec:omics_tokenizer}

\textbf{Composing pathways:} Selecting the appropriate reasoning unit for transcriptomics analysis is challenging, owing to the intricate and hierarchical nature of cellular processes. Pathways, consisting of a group of genes or subpathways involved in a particular biological process, represent a natural reasoning unit for this analysis. A comparison may be drawn to action recognition, where an action (\emph{i.e.,} a biological pathway) can be described by a series of movements captured by sensors (\emph{i.e.,} transcriptomics measurements of a group of genes).



\textbf{Encoding pathways:} Given a set of transcriptomics measurements of $N_{\genes}$ genes, denoted as $\mathbf{g} \in \real^{N_{\genes}}$, and the composition of each pathway, we aim to build pathway-level tokens $\mathbf{X}^{(\pathways)} \in \real^{N_{\pathways} \times d}$, where $d$ denotes the token dimension. Transcriptomics can be seen as tabular data, which can be efficiently encoded with multilayer perceptrons (MLPs). Specifically, we are learning pathway-specific weights $\phi_i$, \emph{i.e.,} $\mathbf{x}_i^{(\pathways)} = \phi_i(\mathbf{g}_{\pathways_i})$, where $\mathbf{g}_{\pathways_i}$ is the gene set present in pathway $\pathways_i$. This can be viewed as learning a \emph{sparse} multi-layer perceptron (S-MLP)~\cite{ma2018using,hao2018pasnet,elmarakeby2021biologically} that maps transcriptomics $\mathbf{g} \in \real^{N_\genes}$ to tokens $\mathbf{x}^{(\pathways)} \in \real^{N_{\pathways}d}$. The network sparsity is controlled by the gene-to-pathway connectivity embedded in the S-MLP weights. By simply reshaping $\mathbf{x}^{(\pathways)} \in \real^{N_{\pathways}d}$ into $\mathbf{X}^{(\pathways)} \in \real^{N_{\pathways} \times d}$, we define pathway tokens that can be used by the Transformer. Each pathway token corresponds to a deep representation of the gene-level transcriptomics that comprises it, which is both (1) interpretable as it encodes a specific biological function and (2) learnable in an end-to-end fashion with respect to the prediction task.

\subsection{Histology Patch Tokenizer from WSIs} \label{sec:histology_tokenizer}

Given an input WSI, we aim to derive low-dimensional patch-level embeddings defining patch tokens. We start by identifying tissue regions to ensure that the background, which carries no biological meaning, is disregarded. Then, we decompose the identified tissue regions into a set of $N_{\histology}$ non-overlapping patches at 20$\times$ magnification (or $\sim0.5\, \mu m$/pixel resolution), that we denote as $\mathbf{H}= \{\mathbf{h}_1, ..., \mathbf{h}_{N_{\histology}}\}$. Due to the large number of patches per WSI (\emph{e.g.,} can be $>$ 50,000 patches or 78 GB as floats), patch embeddings need to be extracted prior to model training to reduce the overall memory requirements. Formally, we employ a pre-trained feature extractor $f(\cdot)$ to map each patch $\mathbf{h}_i$ into a patch embedding as $\mathbf{x}_i^{(\histology)} = f(\mathbf{h}_i)$. In this work, we use a Swin Transformer encoder that was pretrained via contrastive learning on more than 15 million pan-cancer histopathology patches~\cite{wang2022Transformer,wang2021transpath}. The resulting patch embeddings represent compressed representations of the patches (compression ratio of 256), that we further pass through a learnable linear transform to match the token dimension $d$, yielding patch tokens $\mathbf{X}^{(\histology)} \in \real^{N_{\histology} \times d}$. 

\subsection{Multimodal Fusion} \label{sec:multimodal}

We aim to design an early fusion mechanism to model dense multimodal interactions between pathway and patch tokens. We employ Transformer attention~\cite{vaswani2017attention} that measures and aggregates pair-wise interactions between multimodal tokens. Specifically, we define a multimodal sequence by concatenating the pathway and patch tokens resulting in $(N_{\histology}+N_{\pathways})$ tokens of dimensions $d$, and denoted as $\mathbf{X}\in \real^{(N_{\pathways}+N_{\histology}) \times d}$. Following the self-attention terminology~\cite{vaswani2017attention}, we define three linear projections of the tokens using learnable matrices, denoted as $\mathbf{W}_Q \in \real^{d \times d_q}$, $\mathbf{W}_K \in \real^{d \times d_k}$, and $\mathbf{W}_V \in \real^{d \times d_v}$ to extract the queries ($\mathbf{Q}$), keys ($\mathbf{K}$), values ($\mathbf{V}$), and self-attention $\mathbf{A}$, setting $d=d_k=d_q=d_v$. Transformer attention is then defined as:
\begin{align}
    \mathbf{X}_{\text{Att}} = \sigma \Big( \frac{\mathbf{Q}\mathbf{K}^T}{\sqrt{d}} \Big) \mathbf{V} =
    \begin{pmatrix}
        \mathbf{A}_{\pathways \rightarrow\pathways} & \mathbf{A}_{\pathways \rightarrow\histology} \\
        \mathbf{A}_{\histology \rightarrow\pathways} & \mathbf{A}_{\histology \rightarrow\histology} 
    \end{pmatrix}
    \begin{pmatrix}
        \mathbf{V}_{\pathways} \\
        \mathbf{V}_{\histology} 
    \end{pmatrix}
\end{align}
where $\sigma$ is the row-wise softmax. The term $\mathbf{Q}\mathbf{K}^T$ has memory requirements $\mathcal{O}\big((N_{\histology}+N_{\pathways})^2\big)$, which for long sequences becomes expensive to compute.
This constitutes a major bottleneck as a WSI can have $N_{\histology}>50,000$ patches, making this computation challenging on most hardware. Instead, we propose to decompose the multimodal Transformer attention into four intra- and cross-modality terms: (1) the intra-modal pathway self-attention encoding pathway-to-pathway interactions $\mathbf{A}_{\pathways \rightarrow\pathways} \in \real^{N_{\pathways} \times N_{\pathways}}$, (2) the cross-modal pathway-guided cross-attention encoding pathway-to-patch interactions $\mathbf{A}_{\pathways \rightarrow\histology} \in \real^{N_{\pathways} \times N_{\histology}}$, (3) the cross-modal histology-guided cross attention encoding patch-to-pathway interactions $\mathbf{A}_{\histology \rightarrow \pathways} \in \real^{N_{\histology} \times N_{\pathways}}$, and (4) the intra-modal full histology self-attention encoding patch-to-patch interactions $\mathbf{A}_{\histology \rightarrow\histology} \in \real^{N_{\histology} \times N_{\histology}}$. 

As the number of patch tokens is much larger than the number of pathways, \emph{i.e.,} $N_{\histology}>>N_{\pathways}$, most memory requirements come from computing and storing $\textbf{A}_{\histology \rightarrow \histology}$. To address this bottleneck, we approximate Transformer attention as: 
\begin{align} \label{eq:simplified_attention}
    \hat{\mathbf{X}}_{\text{Att}}
    = 
    \begin{pmatrix}
        \mathbf{X}_{\text{Att}}^{(\pathways)} \\
        \hat{\mathbf{X}}_{\text{Att}}^{(\histology)}
    \end{pmatrix}
    = \sigma \left[\frac{1}{\sqrt{d}}
    \begin{pmatrix}
        \mathbf{Q}_{\pathways} \mathbf{K}_{\pathways}^T & \mathbf{Q}_{\pathways}\mathbf{K}_{\histology}^T \\
        \mathbf{Q}_{\histology} \mathbf{K}_{\pathways}^T & -\infty \\
    \end{pmatrix}
    \right]
    \mathbf{V}
\end{align}
where $\mathbf{Q}_{\pathways}$ (respectively $\mathbf{K}_{\pathways}$) and $\mathbf{Q}_{\histology}$ (respectively $\mathbf{K}_{\histology}$) denotes the subset of pathway and histology queries and keys. Setting pre-softmax patch-to-patch interactions to $-\infty$ is equivalent to ignoring these interactions. Expanding Eq.~\ref{eq:simplified_attention}, we obtain that $\mathbf{X}_{\text{Att}}^{(\pathways)} = \sigma \left( \frac{\mathbf{Q}_{\pathways} \mathbf{K}^T}{\sqrt d} \right) \mathbf{V}_{\pathways}$, and $\hat{\mathbf{X}}_{\text{Att}}^{(\histology)} = \sigma \left( \frac{\mathbf{Q}_{\histology} \mathbf{K}_{\pathways}^T}{\sqrt d} \right) \mathbf{V}_{\histology}$. The number of interactions becomes drastically smaller, enabling computing $\hat{\mathbf{A}}$ with limited memory. This formulation can be seen as a sparse attention pattern~\cite{beltagy2020longformer} on a multimodal sequence, where sparsity is imposed between patch tokens. This formulation is parameter-efficient as a unique set of keys, queries, and values is learned for encoding both modalities. Additionally, this formulation resembles a graph neural network on a graph where pathways interconnect, and each pathway links to all patches. After passing $\hat{\mathbf{X}}_{\text{Att}}$ through a feed-forward layer with layer normalization, we take the mean representation of the post-attention pathway and patch tokens denoted as $\bar{\mathbf{x}}_{\text{Att}}^{\pathways}$ and $\bar{\mathbf{x}}_{\text{Att}}^{\histology}$, respectively. The final representation $\bar{\mathbf{x}}_{\text{Att}}$, is then defined by the concatenation of $\bar{\mathbf{x}}_{\text{Att}}^{\pathways}$ and $\bar{\mathbf{x}}_{\text{Att}}^{\histology}$.

\subsection{Survival Prediction} \label{sec:survival}

Using the multimodal embedding $\bar{\mathbf{x}}_{\text{Att}} \in \real^{2d}$, our supervised objective is to predict patient survival. Following previous work~\cite{zadeh2021bias}, we define the patient's survival state by: (1) censorship status $c$, where $c=0$ represents an observed patient death and $c=1$ corresponds to the patient's last known follow-up, and (2) a time-to-event $t_i$, which corresponds to the time between the patient's diagnostic and observed death if $c=0$, or the last follow-up if $c=1$. Instead of directly predicting the observed time of event $t$, we approximate it by defining non-overlapping time intervals $(t_{j-1}, t_j), \;j \in [1, ..., n]$ based on the quartiles of survival time values, and denoted as $y_j$. The problem simplifies to classification, where each patient is now defined by $(\bar{\mathbf{x}}_{\text{Att}}, y_j, c)$. We define our classifier such that each output logit (after sigmoid activation) $\sigma(\hat{y}_j)$ represents the probability that the patient dies during time interval $(t_{j-1}, t_j)$. We further take the cumulative product of the logits as $\prod_{k=1}^j \big(1 - \sigma(\hat{y}_k)\big)$ to represent the probability that the patient survives up to time interval $(t_{j-1}, t_j)$. Finally, by taking the negative of the sum of all logits, we can define a patient-level risk used for training the network. More information are provided in the \textbf{Supplemental}. 

\subsection{Multi-Level Interpretability} \label{sec:multilevel_inter}

We propose an interpretability framework that operates across multiple levels to derive transcriptomics, histology, and cross-modal interpretability (see \textbf{Supplemental}).

\textbf{Transcriptomics:} We employ Integrated Gradient (IG)~\cite{sundararajan2017axiomatic} to identify the influence of \textit{pathways} and \textit{genes}, resulting in a score describing the degree to which each pathway, respectively gene, contributes to predicting the risk. A negative IG score corresponds to a pathway/gene associated with a lower risk, while a positive IG score indicates an association with a higher risk. A very small score denotes negligible influence. Such interpretability analysis serves two purposes: (1) validation of known genes and pathways associated with prognosis and (2) identification of novel gene and pathway candidates that could predict prognosis.

\textbf{Histology:} We process analogously with IG to derive \emph{patch-level} influence that enables studying the morphology of low and high-risk-associated patches.

\textbf{Cross-modal interactions:} Finally, we can study \textit{pathway-to-patch} and \textit{patch-to-pathway} interactions using the learned Transformer attention matrix $\hat{\mathbf{A}}$. Specifically, we define the importance of patch $j$ (respectively pathway) with respect to pathway $i$ (respectively patch) as $\hat{\mathbf{A}}_{ij}$ (respectively $\hat{\mathbf{A}}_{ji}$). This enables building heatmaps correlating a pathway and corresponding morphological features.
This interpretability property is unique to our framework and enables studying how specific cellular functions described by a pathway interact with histology.

\begin{table*}[t]
\caption{Results of $\ours$ and baselines in predicting disease-specific patient survival measured with c-Index (at 20$\times$). Best performance in \textbf{bold}, second best \underline{underlined}. Cat refers to concatenation, KP refers to Kronecker product. All omics and multimodal baselines were trained with the Reactome and Hallmark pathway sets.} 
\label{tab:results20x}
\centering
\resizebox{1.95\columnwidth}{!}{%
\begin{tabular}{ll|cccccc}
\toprule
{} & Model/Study &               BRCA $(\uparrow)$ &               BLCA $(\uparrow)$ &           COADREAD $(\uparrow)$ &               HNSC $(\uparrow)$ &               STAD $(\uparrow)$ &  Overall $(\uparrow)$ \\
\midrule
\parbox[t]{0mm}{\multirow{3}{*}{\rotatebox[origin=c]{90}{WSI}}} & ABMIL~\cite{ilse2018attention} &  $0.493 \small{\pm} 0.126$ &  $0.518 \small{\pm} 0.078$ &  $0.630 \small{\pm} 0.102$ &  $\underline{0.580} \small{\pm} 0.019$ &  $0.550 \small{\pm} 0.077$ &    $0.554$\\

& AMISL~\cite{yao2020whole}                                          &  $0.500 \small{\pm} 0.000$ &  $0.500 \small{\pm}0.000$ &  $0.500 \small{\pm} 0.000$ &  $0.518 \small{\pm} 0.015$ &  $0.506 \small{\pm} 0.014$ &    $0.508$\\
& TransMIL~\cite{shao2021transmil}                                          &  $0.530 \small{\pm} 0.057$ &  $0.551 \small{\pm} 0.091$ &  $0.632 \small{\pm} 0.143$ &  $0.523 \small{\pm} 0.043$ &  $0.544 \small{\pm} 0.080$ &    $0.556$\\
\midrule
\parbox[t]{0mm}{\multirow{3}{*}{\rotatebox[origin=c]{90}{Omics}}} & MLP &  $0.611 \small{\small{\pm}} 0.080$ &   $\underline{0.627} \small{\pm} 0.062$ &   $0.625 \small{\pm} 0.060$ &   $0.548 \small{\pm} 0.045$ &  $\underline{0.586} \small{\pm} 0.098$ &    $\underline{0.599}$\\
& SNN~\cite{klambauer2017self} &  $0.528 \small{\pm} 0.094$ &    $0.584 \small{\pm} 0.113$ &  $0.521 \small{\pm} 0.109$ &  $0.550 \small{\pm} 0.065$ &  $0.565 \small{\pm} 0.080$ &    $0.550$\\
& S-MLP~\cite{elmarakeby2021biologically} &  $0.512 \small{\pm} 0.028$ &  $0.595 \small{\pm} 0.114$ &    $0.581 \small{\pm} 0.066$ &      $0.542 \small{\pm} 0.052$ &  $0.515 \small{\pm} 0.081$ &    $0.549$\\
\midrule
\parbox[t]{0mm}{\multirow{9}{*}{\rotatebox[origin=c]{90}{Multimodal}}} & ABMIL (Cat)~\cite{mobadersany2018predicting} & \multirow{1}{*}{$0.541 \small{\pm} 0.158$} &  \multirow{1}{*}{$0.562 \small{\pm} 0.067$}  & \multirow{1}{*}{$0.592 \small{\pm} 0.102$}  & \multirow{1}{*}{$\underline{0.580} \small{\pm} 0.089$}  & \multirow{1}{*}{$0.523 \small{\pm} 0.098$} &    \multirow{1}{*}{$0.560$} \\ 
& ABMIL (KP)~\cite{chen2022pan} & \multirow{1}{*}{$0.615 \small{\pm} 0.083$} & \multirow{1}{*}{$0.566 \small{\pm} 0.038$} & \multirow{1}{*}{$0.584 \small{\pm} 0.109$} &   \multirow{1}{*}{$0.566 \small{\pm} 0.066$} & \multirow{1}{*}{$0.525 \small{\pm} 0.140$} & \multirow{1}{*}{$0.571$} \\
& AMISL (Cat)~\cite{yao2020whole} & \multirow{1}{*}{$0.462 \small{\pm} 0.179$} &  \multirow{1}{*}{$0.518 \small{\pm} 0.055$} & \multirow{1}{*}{$0.510 \small{\pm} 0.137$}  &  \multirow{1}{*}{$0.478 \small{\pm} 0.051$}  & \multirow{1}{*}{$0.538 \small{\pm} 0.025$}  &    \multirow{1}{*}{$0.501$}  \\
& AMISL (KP)~\cite{yao2020whole} & \multirow{1}{*}{$0.533 \small{\pm} 0.106$} & \multirow{1}{*}{$0.554 \small{\pm} 0.055$} & \multirow{1}{*}{$0.567 \small{\pm} 0.182$}  & \multirow{1}{*}{$0.516 \small{\pm} 0.068$} & \multirow{1}{*}{$0.552 \small{\pm} 0.035$}  &    \multirow{1}{*}{$0.544$} \\

& TransMIL (Cat)~\cite{shao2021transmil} & \multirow{1}{*}{$0.598 \small{\pm} 0.087$} &  \multirow{1}{*}{$\mathbf{0.630} \small{\pm} 0.047$} & \multirow{1}{*}{$0.539 \small{\pm} 0.189$}  &  \multirow{1}{*}{$0.542 \small{\pm} 0.091$}  & \multirow{1}{*}{$0.536 \small{\pm} 0.090$}  &    \multirow{1}{*}{$0.569$}  \\

& TransMIL (KP)~\cite{shao2021transmil} & \multirow{1}{*}{$0.629 \small{\pm} 0.144$} & \multirow{1}{*}{$0.625 \small{\pm} 0.079$} & \multirow{1}{*}{$0.566 \small{\pm} 0.081$}  & \multirow{1}{*}{$0.515 \small{\pm} 0.116$} & \multirow{1}{*}{$0.552 \small{\pm} 0.035$}  &    \multirow{1}{*}{$0.577$} \\

& MOTCat~\cite{xu2023multimodal} &  $ 0.600 \small{\pm} 0.095 $ &  $ 0.596 \small{\pm} 0.079 $ &  $ \underline{0.641} \small{\pm} 0.182 $ &   $0.560 \small{\pm} 0.062$ &   $0.550 \small{\pm} 0.103$ &    $0.589$ \\

& MCAT~\cite{chen2021multimodal} &  $\underline{0.652} \small{\pm} 0.117$ &  $0.598 \small{\pm} 0.094$ &  $0.634 \small{\pm} 0.204$ &   $0.531 \small{\pm} 0.049$ &   $0.557 \small{\pm} 0.101$ &    $0.594$ \\

& $\ours$ (Ours) & $\mathbf{0.655} \small{\pm} 0.089$ &  $0.625 \small{\pm} 0.056$ &  $\mathbf{0.673} \small{\pm} 0.170$ &  $\mathbf{0.600} \small{\pm} 0.061$ &  $\textbf{0.592} \small{\pm} 0.047$ &  $\mathbf{0.629}$ \\

\bottomrule
\end{tabular}
}
\end{table*}

\section{Experiments}

\subsection{Dataset and Implementation}

We evaluate $\ours$ on five datasets from TCGA: Bladder Urothelial Carcinoma (BLCA) (n=359), Breast Invasive Carcinoma (BRCA) (n=869), Stomach Adenocarcinoma (STAD) (n=317), Colon and Rectum Adenocarcinoma (COADREAD) (n=296), and Head and Neck Squamous Cell Carcinoma (HNSC) (n=392). Prior studies have focused on predicting overall survival (OS)~\cite{chen2020pathomic}, however, this approach risks overestimating the proportion of cancer-related deaths as patients may have succumbed to other causes. Instead, we predict disease-specific survival (DSS) as a more accurate representation of the patient's disease status.

\textbf{Pathway collection:} We used the Xena database~\cite{goldman2020visualizing} to access raw transcriptomics from TCGA ($N_{\genes} = 60,499$ in total) along with DSS labels. We extracted pathways from two resources:  Reactome~\cite{gillespie2021reactome} and the Human Molecular Signatures Database (MSigDB) -- Hallmarks~\cite{subramanian2005gene,liberzon2015the}. Reactome and MSigDB--Hallmarks comprise 1,281 and 50 human biological pathways, respectively. We further selected pathways for which at least 90\% of the transcriptomics are available, resulting in 331 pathways derived from 4,999 different genes (281 Reactome pathways from 1,577 genes and 50 Hallmarks pathways from 4,241 genes).

\textbf{Histology collection:} We collected all diagnostic WSIs used for primary diagnosis, resulting in 2,233 WSIs with an average of 14,509 patches per WSI at 20$\times$ (assuming $256 \times 256$ patches). In total, we collected over 2.86 TB of raw image data, comprising around 32.4 million patches. 

\textbf{Implementation:} We used 5-fold cross-validation to train all models. Each split was stratified according to the sample site to mitigate potential batch artifacts~\cite{howard2021the}. To increase variability during training, we randomly sampled 4,096 patches from the WSI. At test time, all patches were used to yield the final prediction (see \textbf{Supplemental}). 

\begin{table*}[t]
\caption{Studying design choices for tokenization (top) and fusion (bottom) in $\ours$ at 20$\times$ magnification. \textbf{Top:} \emph{Single} refers to no tokenization, using tabular transcriptomics features as a single token. \emph{Families} refers to the set of six gene families in MutSigDB, as used in \cite{chen2021multimodal}. \emph{React.+Hallmarks} refers to the main $\ours$~model reported in Table~\ref{tab:results20x}. \textbf{Bottom:} $A_{\pathways \rightarrow \pathways}$ and $A_{\pathways \leftrightarrow \histology}$ refers to pathway-to-pathway, pathway-to-patch, and patch-to-pathway interactions, which is the main $\ours$ model reported in Table~\ref{tab:results20x}. $\mathbf{\tilde{A}}$ refers to using Nyström attention to approximate $\mathbf{A}$.}
\label{tab:ablation20x}
\centering
\resizebox{2.0\columnwidth}{!}{%
\begin{tabular}{ll|cccccc}
\toprule

& Model/Study & BRCA $(\uparrow)$ &               BLCA $(\uparrow)$ &           COADREAD $(\uparrow)$ &               HNSC $(\uparrow)$ &               STAD $(\uparrow)$ &  Overall $(\uparrow)$ \\
\midrule

\parbox[t]{0mm}{\multirow{5}{*}{\rotatebox[origin=c]{90}{Tokenizer}}}  & Single &  $0.625 \small{\pm} 0.149$ &  $0.560 \small{\pm} 0.086$ &  $0.604 \small{\pm} 0.176$ &  $0.580 \small{\pm} 0.075$ &  $0.563 \small{\pm} 0.140$ & $0.586$ \\

& Families &  $0.620 \small{\pm} 0.094$ &  $0.613 \small{\pm} 0.061$ &  $\underline{0.671} \small{\pm} 0.111$ & $\underline{0.600} \small{\pm} 0.076$ & $0.540 \small{\pm} 0.071$ & $0.609$ \\

& Hallmarks & $\underline{0.645} \small{\pm} 0.039$ &  $\mathbf{0.635} \small{\pm} 0.093$ &  $0.633 \small{\pm} 0.151$ &  $0.589 \small{\pm} 0.076$ &  $0.581 \small{\pm} 0.039$ &  $\underline{0.615}$ \\

& Reactome &  $0.579 \small{\pm} 0.044$ &  $0.604 \small{\pm} 0.080$ &  $0.639 \small{\pm} 0.200$ &  $0.574 \small{\pm} 0.061$ &  $\mathbf{0.619} \small{\pm} 0.047$ & ${0.602}$ \\

& React.+Hallmarks &  \multirow{1}{*}{$\mathbf{0.655} \small{\pm} 0.089$} &  \multirow{1}{*}{$\underline{0.625} \small{\pm} 0.056$} &  \multirow{1}{*}{$\mathbf{0.673} \small{\pm} 0.170$} &  \multirow{1}{*}{$\mathbf{0.600} \small{\pm} 0.061$} &  \multirow{1}{*}{$0.592 \small{\pm} 0.047$} &    \multirow{1}{*}{$\mathbf{0.629}$} \\

\midrule
\parbox[t]{0mm}{\multirow{4}{*}{\rotatebox[origin=c]{90}{Fusion}}} & $\mathbf{A}_{\pathways \rightarrow \pathways}$, $\mathbf{A}_{\pathways \rightarrow \histology}$ &   $0.446\small{\pm} 0.116$ &  $0.603 \small{\pm} 0.038$ &  $0.565 \small{\pm} 0.166$ &  $0.526 \small{\pm} 0.030$ &  $0.582 \small{\pm} 0.053$ &    $0.544$ \\
& $\mathbf{A}_{\pathways \rightarrow \pathways}$, $\mathbf{A}_{\histology \rightarrow \pathways}$ &  $0.546 \small{\pm} 0.118$ &  $0.589 \small{\pm} 0.037$ &  $0.633 \small{\pm} 0.130$ &   $0.498 \small{\pm} 0.037$ &  $0.480 \small{\pm} 0.083$ &    $0.549$ \\
& $\mathbf{A}_{\pathways \rightarrow \pathways}$, $\mathbf{A}_{\histology \rightarrow \pathways}$, $\mathbf{A}_{\pathways \rightarrow \histology}$ &  \multirow{1}{*}{$\mathbf{0.655} \small{\pm} 0.089$} &  \multirow{1}{*}{$\underline{0.625} \small{\pm} 0.056$} &  \multirow{1}{*}{$\mathbf{0.673} \small{\pm} 0.170$} &  \multirow{1}{*}{$\mathbf{0.600} \small{\pm} 0.061$} &  \multirow{1}{*}{$0.592 \small{\pm} 0.047$} &    \multirow{1}{*}{$\mathbf{0.629}$} \\
& $\mathbf{\tilde{A}}$~\cite{xiong2021nystrom} &  $0.555 \small{\pm} 0.066$ &  $0.565 \small{\pm} 0.101$ &  $0.612 \small{\pm} 0.194$ &   $0.508 \small{\pm} 0.032$ &   $\underline{0.493} \small{\pm} 0.086$ &    $0.547$ \\
\bottomrule
\end{tabular}
}
\end{table*}

\subsection{Baselines}

We group as: (1) unimodal histology methods, (2) unimodal transcriptomics methods, and (3) multimodal methods (further sub-categorized into early \emph{vs.} late fusion methods).

\textbf{Histology baselines:} All baselines use the same pre-trained feature extractor as $\ours$ based on~\cite{wang2021transpath}. We compare with \textit{ABMIL}~\cite{ilse2018attention}, which uses a gated-attention pooling, \textit{AMISL}~\cite{yao2020whole}, which first clusters patch embeddings using K-means before attention, and \textit{TransMIL}~\cite{shao2021transmil}, that approximates patch self-attention with Nyström method~\cite{xiong2021nystrom}. 

\textbf{Transcriptomics baselines:} All baselines use the same input defined by aggregating Reactome and Hallmarks transcriptomics. (a) \textit{MLP}~\cite{haykin1994neural} uses a 4-layer MLP, (b) \textit{SNN}~\cite{haykin1994neural,chen2020pathomic} supplements \textit{MLP} with additional alpha dropout layers, and (c) \textit{S-MLP}~\cite{hao2018pasnet,elmarakeby2021biologically} uses a 2-layer sparse pathway-aware MLP followed by a dense 2-layer MLP. This baseline shares similarities with our transcriptomics encoder.

\textbf{Multimodal baselines:} (a) \textbf{Late fusion:} We combine ABMIL~\cite{ilse2018attention}, AMISL~\cite{yao2020whole}, and TransMIL~\cite{shao2021transmil} with an S-MLP using concatenation~\cite{mobadersany2018predicting}, denoted as \textit{ABMIL (Cat)}, \textit{AMISL (Cat)}, and \textit{TransMIL (Cat)}, and Kronecker product~\cite{chen2020pathomic, fukui2016multimodal, zadeh2017tensor, weng2019multimodal}, denoted as \textit{ABMIL (KP)}, \textit{AMISL (KP)}, and \textit{TransMIL (KP)}. (b) \textbf{Early fusion:} \textit{MCAT}~\cite{chen2021multimodal} which uses genomic-guided cross-attention followed by modality-specific self-attention blocks, and \textit{MOTCat}~\cite{xu2023multimodal} which uses Optimal Transport (OT) for matching the patch token and genomic token distributions. 

\subsection{Survival Prediction Results}

Table~\ref{tab:results20x} present results of $\ours$ and baselines evaluated at 20$\times$ magnification (see \textbf{Supplemental} for 10$\times$ analysis). $\ours$ reaches best overall performance, outperforming unimodal and multimodal baselines at both 20$\times$ and 10$\times$. At 20$\times$, $\ours$ reaches +7.3$\%$ compared to TransMIL, $+3.0\%$ compared to MLP, and $+3.5$ compared to MCAT. We attribute the high performance of $\ours$ to (1) the use of both modalities, (2) a unified, simple, and parameter-efficient fusion model, and (3) a semantically meaningful transcriptomics tokenizer. 

\textbf{Transcriptomics \emph{vs}. Histology \emph{vs.} Multimodal:} Multimodal baselines significantly outperform histology baselines. Interestingly, a simple MLP trained on our set of transcriptomics constitutes a strong baseline that outperforms several multimodal methods. This highlights the challenge of performing robust feature selection and integrating heterogeneous and high-dimensional data modalities. In addition, the relatively small dataset size further complicates the learning of complex models and risk over-fitting. Comparisons against clinical variables are provided in \textbf{Supplemental}. 

\textbf{Context \emph{vs.} No context:} ABMIL and TransMIL perform similarly despite TransMIL modeling path-to-patch interactions using Nyström attention. This observation supports our design choice of disregarding patch-to-patch interactions. In addition, $\ours$ performance is similar across magnifications ($0.629$ overall c-index in both cases). This observation also holds for most histology and multimodal baselines. 

\textbf{Sparse \emph{vs.} dense transcriptomics encoders:} A dense MLP yields better performance than a sparse pathway-aware MLP. However, sparse networks have shown to be particularly parameter-efficient when the number of genes considered drastically increases and are more interpretable than regular MLPs~\cite{elmarakeby2021biologically}. As the number of genes increases, this trend might evolve. 

\textbf{Early \emph{vs.} Late fusion:} \emph{Early fusion} methods (MCAT~\cite{chen2021multimodal}, MOTCat~\cite{xu2023multimodal} and $\ours$) outperform all late fusion methods. We attribute this observation to the creation of a joint feature space that can model fined-grained interactions between transcriptomics and histology tokens. Overall, these findings justify the need for (1) modeling dense interactions between pathway and patch tokens and (2) unifying fusion in a single Transformer attention.

\subsection{Ablation Study}

To evaluate our design choices, we performed a series of ablations studying different \emph{Tokenizers} and \emph{Fusion} schemes.

\textbf{Tokenizer:} $\ours$ employs the Reactome and Hallmarks databases as sources of biological pathways. We assess the model performance when using each database in isolation, as well as using all genes assigned to one token (\emph{Single}) and the gene families used in~\cite{chen2021multimodal}. 
With increased granularity of transcriptomics tokens, the overall performance increases, showing that building semantic tokens brings interpretability properties and improves performance. We attribute this observation to the fact that each token encodes more and more specific biological functions, enabling better cross-modal modeling.


\textbf{Fusion:} We ablate $\ours$ by further simplifying Transformer attention to its left part considering $A_{\pathways \rightarrow \pathways}$ and $A_{\histology \rightarrow \pathways}$, and to its top part $A_{\pathways \rightarrow \pathways}$ and $A_{\pathways \rightarrow \histology}$ (this design resembles MCAT~\cite{chen2021multimodal} where a single, shared multimodal attention layer is learned).
Both branches bring complementary information (observed decrease of $-5.6\%$ and $-7.5\%$ in c-index), justifying the need to model both pathway-to-patch and patch-to-pathways interactions. We further adapt $\ours$ with Nyström attention that enables training on very long sequences by simplifying self-attention with a low-rank approximation. This yields significantly worse performance $-6.9\%$. We hypothesize that the ``true full attention" has low-entropy, making it more challenging to be approximated by low-rank methods~\cite{chen2021scatterbrain}, and that sparse attention patterns offer better approximations.

\begin{figure*}[t]
    \centering
    \includegraphics[width=0.95\textwidth]{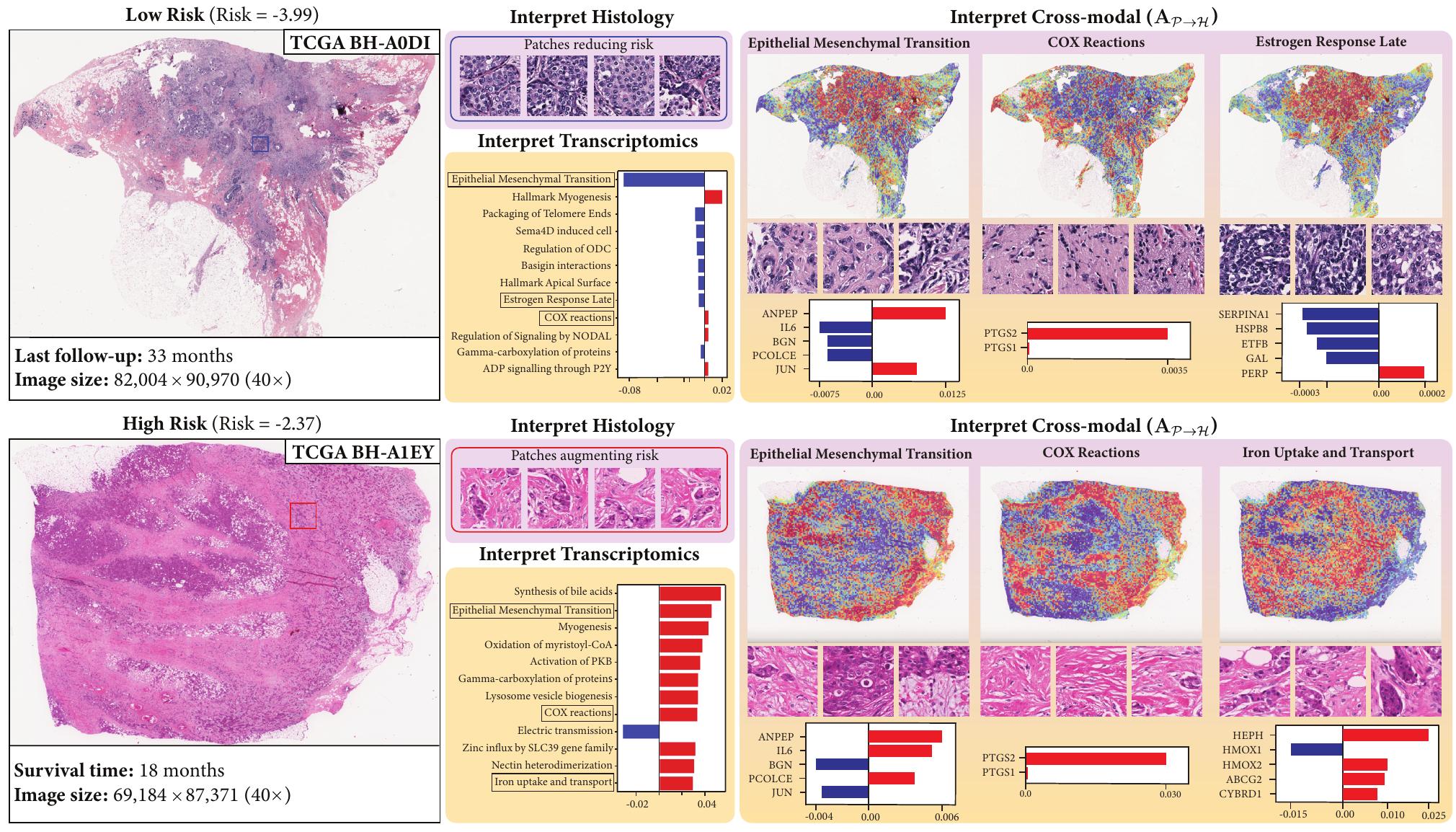}
    \caption{\textbf{Multi-level interpretability visualization in a breast cancer patient.} \textbf{Top:} Low-risk patient. \textbf{Bottom:} High-risk patient. Genes and pathways in red increase risk, and those in blue decrease risk. Heatmap colors indicate importance, with red indicating high importance and blue indicating low importance. The pathways and morphologies identified as important in these cases generally correspond well with patterns that have been previously described in invasive breast cancer (\emph{e.g.} Estrogen Response Late).} 
    \label{fig:interpretability_brca}
\end{figure*}

\subsection{Interpretability}

Examination of the multi-level interpretability can lead to novel biological insight regarding the interplay between pathways and histology in determining a patient's risk. Here, we compare a low (top) and high (bottom) risk case of breast invasive carcinoma (BRCA) (Fig.~\ref{fig:interpretability_brca}) and bladder urothelial carcinoma (BLCA) (see \textbf{Supplemental}).

In analyzing Fig.~\ref{fig:interpretability_brca}, we observe that several pathways have high absolute importance scores in the low and high-risk cases, most notably the Hallmark Epithelial-Mesenchymal Transition (EMT)~\cite{wang2013epithelialmesenchymal} and COX Reactions pathways~\cite{mazhar2006cox}, both of which are known to be involved in breast cancer. EMT is thought to underlie tumor cells' ability to invade and metastasize~\cite{kalluri2009basics}, and the inverse importance of this pathway for the low- and high-risk cases is compatible with this analysis. This finding is enforced by studying the cross-modal interpretability that highlights the association of EMT with nests of tumor cells invading stroma. Members of the COX family of cyclooxygenases, especially COX-2, have also been implicated in breast carcinogenesis and are being investigated as a component of therapeutic regimens~\cite{harris2014cyclooxygenase}. Cross-modal interpretability demonstrates stromal and immune cells in both cases. Though there is some overlap between important pathways in the two cases in Fig.~\ref{fig:interpretability_brca}, the majority differ between the two. For instance, in the high-risk case, a pathway relating to iron metabolism (a known contributor to breast carcinogenesis and prognosis~\cite{torti2013cellular}) was identified, with patches showing small nests of tumor cells invading through a dense stroma. In the low-risk case, a pathway relating to the cellular response to estrogen was found to be important, with corresponding patches demonstrating lower-grade invasive carcinoma or carcinoma in situ morphologies, consistent with others' observation that hormone-positive breast cancers tend to be lower grade and have longer survival times~\cite{dunnwald2007hormone}. 
Interestingly, the Hallmark Myogenesis pathway is assigned relatively high positive importance for both cases in Fig.~\ref{fig:interpretability_brca}. Myogenesis has not been extensively studied in breast cancer, but it is plausible that tumor cells either themselves express genes involved in this pathway as part of their epithelial-mesenchymal transition or they induce stromal cells to do so. This highlights the ability of our method to drive novel biological insight for subsequent investigation.


The flexibility of our approach in providing unimodal and cross-modal interpretability allows us to uncover novel multimodal biomarkers of prognosis that could conceivably be used to design better cancer therapies. As our understanding of the molecular underpinnings of disease grows, the interpretability of $\ours$ may spur research into the possibility of targeting specific combinations of morphologies and pathways. 

\section{Conclusion}

This paper addresses two challenges posed by the multimodal fusion of transcriptomics and histology: (1) we address the challenge of transcriptomics tokenization by defining \emph{biological pathway} tokens that encode semantically meaningful and interpretable functions, and (2) we overcome the computational challenge of integrating long multimodal sequences by designing a multimodal Transformer with sparse modality-specific attention patterns. Our model achieves state-of-the-art survival performance when tested on five datasets from TCGA. In addition, our interpretability framework reveals known and candidate prognostic features. While our interpretability framework enables identifying prognostic features, these findings remain qualitative. Future work could focus on interpretability metrics that generalize findings at dataset-level, \emph{e.g.,} with quantitative morphological characterizations of specific pathways. In addition, our findings suggest that including patch-to-patch interactions does not lead to improved performance. Nonetheless, the absence of a performance boost should not be an evidence that patch-to-patch interactions are unnecessary, but rather that modeling such interactions is a challenging problem that remains to be solved.



{
    \small
    \bibliographystyle{ieeenat_fullname}
    \bibliography{main}
}

\clearpage
\setcounter{page}{1}
\setcounter{section}{0}
\setcounter{figure}{0}
\setcounter{table}{0}
\setcounter{equation}{0}
\maketitlesupplementary

\section{Survival prediction}

Following the notation introduced previously, we aim to predict patient survival from the multimodal embedding $\bar{\mathbf{x}}_{\text{Att}} \in \real^{2d}$. Consistently with previous work~\cite{zadeh2021bias}, we define the patient's survival state by: (1) censorship status $c$, where $c=0$ represents an observed patient death and $c=1$ corresponds to the patient's last known follow-up, and (2) a time-to-event $t_i$, which corresponds to the time between the patient's diagnostic and observed death if $c=0$, or the last follow-up if $c=1$. Instead of directly predicting the observed time of event $t$, we approximate it by defining non-overlapping time intervals $(t_{j-1}, t_j), \;j \in [1, ..., n]$ based on the quartiles of survival time values, and denoted as $y_j$. The problem simplifies to classification with censorship information, where each patient is now defined by $(\bar{\mathbf{x}}_{\text{Att}}, y_j, c)$. We build a classifier such that each output logit predicted by the network $\hat{y}_j$ correspond to a time interval. From there, we define the discrete hazard function $f_{\text{hazard}}(y_j | \bar{\mathbf{x}}_{\text{Att}}) = S(\hat{y}_j)$ where $S$ is the sigmoid activation. Intuitively, $f_{\text{hazard}}(y_j | \bar{\mathbf{x}}_{\text{Att}})$ represents the probability that the patient dies during time interval $(t_{j-1}, t_j)$. Additionally, we define the discrete survival function $f_{\text{surv}}(y_j | \bar{\mathbf{x}}_{\text{Att}}) = \prod_{k=1}^j \big(1 - f_{\text{hazard}}(y_k | \bar{\mathbf{x}}_{\text{Att}})\big)$ that represents the probability that the patient survives up to time interval $(t_{j-1}, t_j)$. These enable us to define the negative log-likelihood (NLL) survival loss~\cite{zadeh2021bias}, which generalizes NLL to data with censorship. Formally, we express it as:
\begin{align}\label{eq:rank_loss}
    \mathcal{L}\Big(&\{\bar{\mathbf{x}}^{(i)}_{\text{Att}}, y^{(i)}_j, c^{(i)} \}_{i=1}^{N_{\dataset}} \Big) =\\
    &\sum_{i=1}^{N_{\dataset}}
    -c^{(i)} \log(f_{\text{surv}}(y_j^{(i)} | \bar{\mathbf{x}}^{(i)}_{\text{Att}})) \label{eq:nnl_1} \\ 
    &+ (1-c^{(i)}) \log(f_{\text{surv}}(y_j^{(i)} -1 | \bar{\mathbf{x}}^{(i)}_{\text{Att}})) \label{eq:nnl_2} \\ 
    &+ (1-c^{(i)}) \log(f_{\text{hazard}}(y_j^{(i)} | \bar{\mathbf{x}}^{(i)}_{\text{Att}})) \label{eq:nnl_3}
\end{align}
where $N_{\dataset}$ is the number of samples in the dataset. Intuitively, Eq.~\ref{eq:nnl_1} enforces a high survival probability for patients who remain alive after the final follow-up, Eq.~\ref{eq:nnl_2} enforces that patients that died have high survival up to the time stamp where death was observed, and Eq.~\ref{eq:nnl_3} ensures that the correct timestamp is predicted for patients for whom death is observed. A thorough mathematical description can be found in~\cite{zadeh2021bias}. 

Finally, by taking the negative of the sum of all logits, we can define a patient-level risk used to identify different risk groups and stratify patients.

\section{Implementation}

\subsection{Model training}

The code was implemented using Python 3.9, models were implemented in PyTorch and the interpretability was based on Captum~\cite{kokhlikyan2020captum}. $\ours$, baselines and ablations were optimized using the RAdam optimizer~\cite{liu2019variance}, a batch size of 1, a learning rate of $5 \times 10^{-4}$, and $10^{-3}$ weight decay. The patch encoder yields 768-dimensional embeddings (CTransPath output) that are projected to $d=256$, the token dimension. The transcriptomics encoder is composed of 2-layer feed-forward networks with alpha dropout~\cite{klambauer2017self} to yield pathway tokens. The Transformer is implemented with a single head and layer, without class (CLS) token. The transformer is followed by a layernorm, a feed-forward layer, and a 2-layer classification head. All model training was done using a single NVIDIA RTX 3090Ti.

\subsection{Metrics}

The models are evaluated using (1) the concordance index (c-index, higher is better), which measures the proportion of all possible pairs of observations where the model's predicted values correctly predict the ordering of the actual survival (ranges from 0.5 (random prediction) to 1.0 (perfect prediction)), and (2) Kaplan-Meier (KM) curves that enable visualizing the probability of survival of patients of different risk groups over a certain period of time. We apply the logrank statistical significance test to determine if the separation between low and high-risk groups is statistically significant $(\text{p-value} < 0.05)$. 

\section{Additional interpretability}

\begin{figure}[t]
    \centering
    \includegraphics[width=0.44\textwidth]{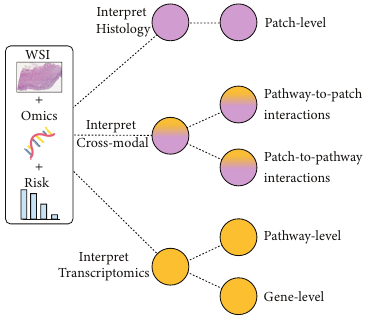}
    \caption{\textbf{Multi-level interpretability framework.} From the multimodal input consisting of a WSI and transcriptomic measurements, and the predicted risk, we can attribute risk at slide-, gene- and biological pathway-level. The framework also enables studying pathway-to-patch interactions and patch-to-pathway interactions for unravelling correspondences between the two modalities.}
    \label{fig:inter_overview}
\end{figure}

A high-level depiction of the proposed multi-level interpretability framework is shown in Fig.~\ref{fig:inter_overview}. 

\begin{figure*}[t]
    \centering
    \includegraphics[width=0.99\textwidth]{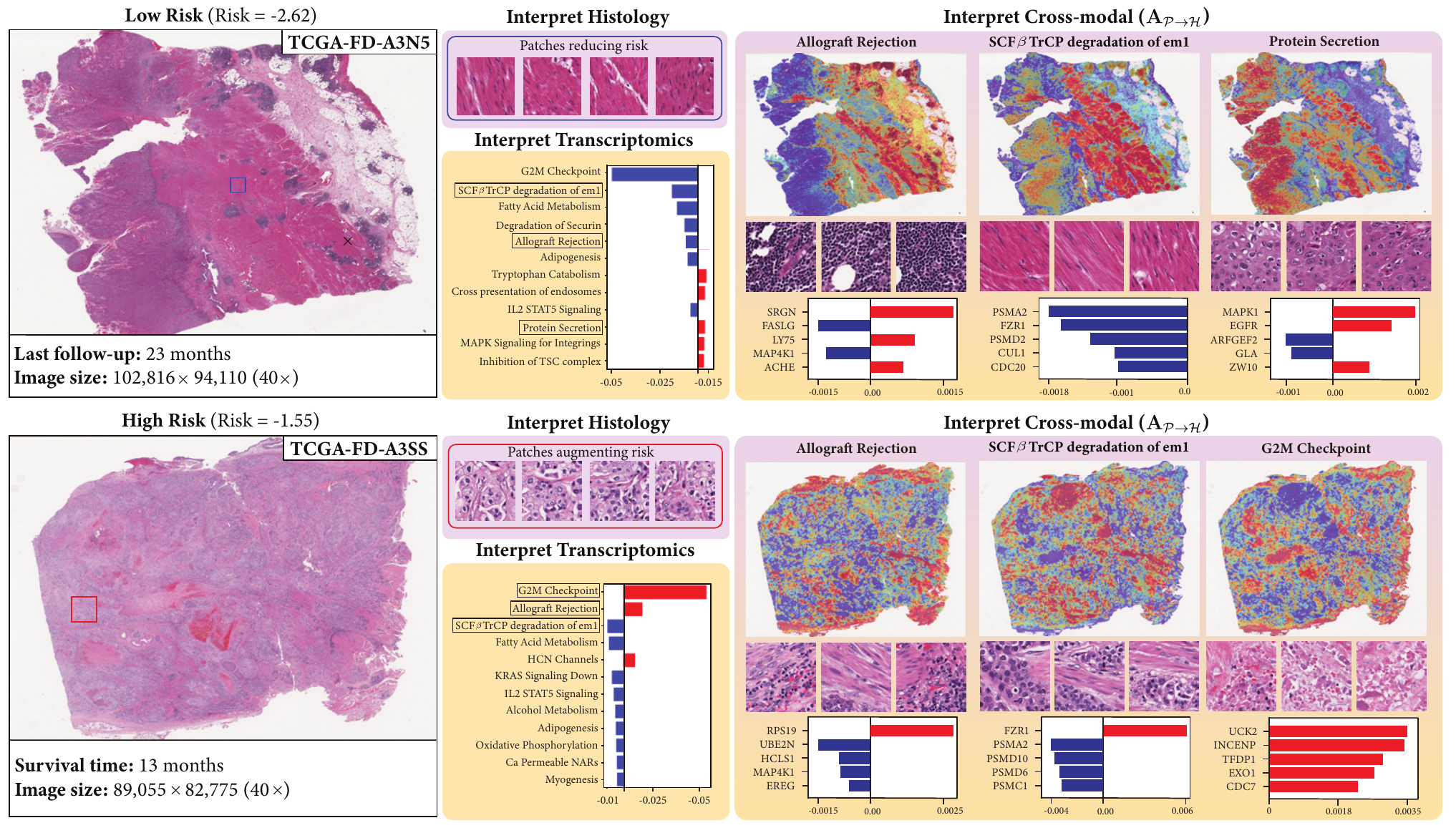}
    \caption{\textbf{Multi-level interpretability visualization in a bladder cancer patient.} \textbf{Top:} Low-risk patient. \textbf{Bottom:} High-risk patient. Genes and pathways in red increase risk, and those in blue decrease risk. Heatmap colors indicate importance, with red indicating high importance and blue indicating low importance. The pathways and morphologies identified as important in these cases generally correspond well with patterns that have been previously described in bladder urothelial carcinoma (e.g., the G2M checkpoint).}
    \label{fig:interpretability_blca}
\end{figure*}

To complement the interpretability analysis presented in the main paper, we further analyze a low and high risk case from \textbf{BLCA} (see Fig~\ref{fig:interpretability_blca}). The histology interpretation indicates that the presence of healthy bladder muscle reduces risk, and pleomorphic tumor cells with foamy cytoplasm contribute to augmenting risk. The majority of important pathways relate to cell cycle control (e.g., G2M checkpoint, SCF $\beta$ TrCP degradation of em1), metabolism (e.g., fatty acid metabolism), and immune-related function (allograft rejection and IL2 STAT5 signaling). The contributions of pathways to overall risk are also in line with previous literature. For example, previous pathway expression analyses have found G2M checkpoint and immune-related pathways to be significant in predicting bladder cancer prognosis~\cite{jiang2021identification}. Qualitative assessments of the cross-modal interactions found by $\ours$ are scientifically plausible. For example, the allograft rejection pathway consists of multiple genes that are activated in immune response to allografts and cancer. In the low-risk case, allograft rejection highly attends to tumor-infiltrating lymphocytes and collections of lymphocytes within and near the muscular wall of the bladder. In the higher-risk case, this pathway again attends to collections of inflammatory cells that are interspersed within the muscular wall. The SCF $\beta$ TrCP degradation of em1 pathway is important in controlling cell division by mitosis. In the low-risk case, this pathway attends to uninvolved bladder muscle, whereas in the high-risk case, the same pathway attends to tumor cells invading the bladder muscle. While there is an overlap between pathways for low and high-risk cases, $\ours$ also identifies pathways present in only one case. For example, in the low-risk case, $\ours$ finds the protein secretion pathway to be highly attending to tumor cells and not the healthy bladder muscle cells. In both cases, the G2M checkpoint pathway (critical for the healthy progress of the cell cycle) is found to be important. In the high-risk case, we see this pathway contributing largely to increasing risk. Interestingly, we also find that this pathway attends to large areas of necrosis, which is reasonable given that aberrations in cell cycle regulation lead to cell death. 

\section{Additional results}

\textbf{10$\times$ results:} We also present an analysis of $\ours$ and baselines (Tab.~\ref{tab:results10x}), and ablations (Tab.~\ref{tab:ablation_10x}) at 10$\times$ magnification. Trends from the 20$\times$ analysis remain in that (1) $\ours$ achieves overall best performance, (2) transcriptomic baselines remain strong competitors, and (3) multimodal models provide better overall performance. Interestingly, $\ours$ at 10$\times$ and 20$\times$ provide the same performance (62.9\% over the five cohorts).  

\begin{table*}[t]
\caption{Results of $\ours$ and baselines in predicting disease-specific patient survival measured with c-Index (at 10x). Best performance in \textbf{bold}, second best \underline{underlined}. Cat refers to concatenation, KP refers to Kronecker product. All omics and multimodal baselines were trained with the Reactome and Hallmark pathway sets. The omics baselines are carried forward from the 20x experiments.} 
\label{tab:results10x}
\centering
\resizebox{1.95\columnwidth}{!}{%
\begin{tabular}{ll|cccccc}
\toprule
{} & Model/Study &               BRCA $(\uparrow)$ &               BLCA $(\uparrow)$ &           COADREAD $(\uparrow)$ &               HNSC $(\uparrow)$ &               STAD $(\uparrow)$ &  Overall $(\uparrow)$ \\

\midrule
\parbox[t]{0mm}{\multirow{3}{*}{\rotatebox[origin=c]{90}{WSI}}} & ABMIL~\cite{ilse2018attention} 
&  $0.604$$ \small{\pm} 0.110$ &  $0.518 \small{\pm} 0.078$ &  $0.652 \small{\pm} 0.192$ &  $0.572 \small{\pm} 0.070$ &  $0.522 \small{\pm} 0.136$ &    $0.574$\\

& AMISL~\cite{yao2020whole}                                          
&  $0.500 \small{\pm} 0.000$ &  $0.500 \small{\pm}0.000$ &  $0.506 \small{\pm} 0.012$ &  $0.498 \small{\pm} 0.050$ &  $0.500 \small{\pm} 0.000$ &    $0.501$\\

& TransMIL~\cite{shao2021transmil}                                          
&  $0.527 \small{\pm} 0.157$ &  $0.541 \small{\pm} 0.043$ &  $0.628 \small{\pm} 0.193$ &  $0.557 \small{\pm} 0.056$ &  $0.516 \small{\pm} 0.080$ &    $0.554$\\
\midrule

\parbox[t]{0mm}{\multirow{3}{*}{\rotatebox[origin=c]{90}{Omics}}} & MLP &  $0.611 \small{\small{\pm}} 0.080$ &   $\underline{0.627} \small{\pm} 0.062$ &   $0.625 \small{\pm} 0.060$ &   $0.548 \small{\pm} 0.045$ &  $0.586 \small{\pm} 0.098$ &    $\underline{0.599}$\\

& SNN~\cite{klambauer2017self} &  $0.528 \small{\pm} 0.094$ &    $0.584 \small{\pm} 0.113$ &  $0.521 \small{\pm} 0.109$ &  $0.550 \small{\pm} 0.065$ &  $0.565 \small{\pm} 0.080$ &    $0.550$\\

& S-MLP~\cite{elmarakeby2021biologically} &  $0.512 \small{\pm} 0.028$ &  $0.595 \small{\pm} 0.114$ &    $0.581 \small{\pm} 0.066$ &      $0.542 \small{\pm} 0.052$ &  $0.515 \small{\pm} 0.081$ &    $0.549$\\

\midrule
\parbox[t]{0mm}{\multirow{8}{*}{\rotatebox[origin=c]{90}{Multimodal}}} & ABMIL (Cat)~\cite{mobadersany2018predicting} & \multirow{1}{*}{$\underline{0.623} \small{\pm} 0.066$} &  \multirow{1}{*}{$0.619 \small{\pm} 0.094$}  & \multirow{1}{*}{$0.622 \small{\pm} 0.165$}  & \multirow{1}{*}{$0.549 \small{\pm} 0.063$}  & \multirow{1}{*}{$0.547 \small{\pm} 0.111$} &    \multirow{1}{*}{$0.592$} \\ 

& ABMIL (KP)~\cite{chen2022pan} & \multirow{1}{*}{$0.529 \small{\pm} 0.099$} & \multirow{1}{*}{$0.592 \small{\pm} 0.086$} & \multirow{1}{*}{$0.640 \small{\pm} 0.183$} &   \multirow{1}{*}{$\underline{0.596} \small{\pm} 0.039$} & \multirow{1}{*}{$0.526 \small{\pm} 0.107$} & \multirow{1}{*}{$0.577$} \\

& AMISL (Cat)~\cite{yao2020whole} & \multirow{1}{*}{$0.508 \small{\pm} 0.131$} &  \multirow{1}{*}{$0.543 \small{\pm} 0.069$} & \multirow{1}{*}{$0.620 \small{\pm} 0.110$}  &  \multirow{1}{*}{$0.539 \small{\pm} 0.051$}  & \multirow{1}{*}{$0.583 \small{\pm} 0.104$}  &    \multirow{1}{*}{$0.559$}  \\

& AMISL (KP)~\cite{yao2020whole} & \multirow{1}{*}{$0.551 \small{\pm} 0.122$} & \multirow{1}{*}{$0.500 \small{\pm} 0.068$} & \multirow{1}{*}{$0.518 \small{\pm} 0.151$}  & \multirow{1}{*}{$0.523 \small{\pm} 0.063$} & \multirow{1}{*}{$0.565 \small{\pm} 0.062$}  &    \multirow{1}{*}{$0.531$} \\

& TransMIL (Cat)~\cite{shao2021transmil} & \multirow{1}{*}{$0.539 \small{\pm} 0.072$} &  \multirow{1}{*}{$0.598 \small{\pm} 0.043$} & \multirow{1}{*}{$0.632 \small{\pm} 0.200$}  &  \multirow{1}{*}{$0.537 \small{\pm} 0.065$}  & \multirow{1}{*}{$0.547 \small{\pm} 0.094$}  &    \multirow{1}{*}{$0.571$}  \\

& TransMIL (KP)~\cite{shao2021transmil} & \multirow{1}{*}{$0.538 \small{\pm} 0.054$} & \multirow{1}{*}{$0.603 \small{\pm} 0.043$} & \multirow{1}{*}{$\mathbf{0.686} \small{\pm} 0.195$}  & \multirow{1}{*}{$0.521 \small{\pm} 0.111$} & \multirow{1}{*}{$0.459 \small{\pm} 0.170$}  &    \multirow{1}{*}{$0.561$} \\

& MOTCat~\cite{xu2023multimodal} &  $0.612 \small{\pm} 0.156$ &  $0.614 \small{\pm} 0.079$ &  $0.569 \small{\pm} 0.191$ &   $0.592 \small{\pm} 0.080$ &   $\underline{0.586} \small{\pm} 0.056$ &    $0.595$ \\

& MCAT~\cite{chen2021multimodal} &  $0.473 \small{\pm} 0.123$ &  $0.545 \small{\pm} 0.070$ &  $0.480 \small{\pm} 0.243$ &   $0.494 \small{\pm} 0.072$ &   $0.433 \small{\pm} 0.064$ &    $0.485$ \\

& $\ours$ (Ours) & $\mathbf{0.640} \small{\pm} 0.093$ &  $\mathbf{0.628} \small{\pm} 0.073$ &  $\underline{0.675} \small{\pm} 0.175$ &  $\mathbf{0.605} \small{\pm} 0.068$ &  $\mathbf{0.598} \small{\pm} 0.081$ &  $\mathbf{0.629}$ \\

\bottomrule
\end{tabular}
}
\end{table*}

\begin{table*}[t]
\caption{Studying design choices for tokenization (top) and fusion (bottom) in \ours at 10$\times$ magnification. \textbf{Top:} \emph{Single} refers to no tokenization, using tabular transcriptomics features as a single token. \emph{Families} refers to the set of six gene families in MutSigDB, as used in \cite{chen2021multimodal}. \emph{React.+Hallmarks} refers to the main \ours~model reported in Table~\ref{tab:results10x}. \textbf{Bottom:} $A_{\pathways \rightarrow \pathways}$ and $A_{\pathways \leftrightarrow \histology}$ refers to pathway-to-pathway, pathway-to-patch, and patch-to-pathway interactions, which is the main $\ours$ model reported in Table~\ref{tab:results10x}. $\mathbf{\tilde{A}}$ refers to using Nyström attention to approximate $\mathbf{A}$.}
\label{tab:ablation_10x}
\centering
\resizebox{1.95\columnwidth}{!}{%
\begin{tabular}{ll|cccccc}
\toprule

& Model/Study & BRCA $(\uparrow)$ &               BLCA $(\uparrow)$ &           COADREAD $(\uparrow)$ &               HNSC $(\uparrow)$ &               STAD $(\uparrow)$ &  Overall $(\uparrow)$ \\
\midrule

\parbox[t]{0mm}{\multirow{5}{*}{\rotatebox[origin=c]{90}{Tokenizer}}}  & Single &  $0.617 \small{\pm} 0.147$ &  $0.599 \small{\pm} 0.077$ &  $0.533 \small{\pm} 0.07$ &  $0.544 \small{\pm} 0.077$ &  $0.524 \small{\pm} 0.117$ & $0.563$ \\

& Families &  $0.534 \small{\pm} 0.156$ &  $0.588 \small{\pm} 0.060$ &  $\mathbf{0.686} \small{\pm} 0.156$ & $0.543 \small{\pm} 0.077$ & $0.457 \small{\pm} 0.077$ & $0.562$ \\

& Hallmarks & $0.609 \small{\pm} 0.087$ &  $\underline{0.633} \small{\pm} 0.090$ &  $0.659 \small{\pm} 0.117$ &  $0.601 \small{\pm} 0.031$ &  $0.580 \small{\pm} 0.052$ &  ${0.616}$ \\

& Reactome &  $\mathbf{0.665} \small{\pm} 0.086$ &  $\mathbf{0.634} \small{\pm} 0.077$ &  $0.626 \small{\pm} 0.157$ &  $\mathbf{0.611} \small{\pm} 0.067$ &  $\mathbf{0.603} \small{\pm} 0.033$ & $\underline{0.628}$ \\

& React.+Hallmarks &  \multirow{1}{*}{$\underline{0.640} \small{\pm} 0.093$} &  \multirow{1}{*}{$0.628 \small{\pm} 0.073$} &  \multirow{1}{*}{$\underline{0.675} \small{\pm} 0.175$} &  \multirow{1}{*}{$\underline{0.605} \small{\pm} 0.068$} &  \multirow{1}{*}{$0.598 \small{\pm} 0.081$} &    \multirow{1}{*}{$\mathbf{0.629}$} \\

\midrule
\parbox[t]{0mm}{\multirow{4}{*}{\rotatebox[origin=c]{90}{Fusion}}} & $\mathbf{A}_{\pathways \rightarrow \pathways}$, $\mathbf{A}_{\pathways \rightarrow \histology}$ &   $0.589\small{\pm} 0.077$ &  $0.570 \small{\pm} 0.099$ &  $0.594 \small{\pm} 0.124$ &  $0.568 \small{\pm} 0.067$ &  $0.546 \small{\pm} 0.135$ &    $0.573$ \\

& $\mathbf{A}_{\pathways \rightarrow \pathways}$, $\mathbf{A}_{\histology \rightarrow \pathways}$ &  $0.573 \small{\pm} 0.085$ &  $0.577 \small{\pm} 0.118$ &  $0.531 \small{\pm} 0.221$ &   $0.566 \small{\pm} 0.064$ &  $0.521 \small{\pm} 0.056$ &    $0.554$ \\

& $\mathbf{A}_{\pathways \rightarrow \pathways}$, $\mathbf{A}_{\histology \rightarrow \pathways}$, $\mathbf{A}_{\pathways \rightarrow \histology}$ &  \multirow{1}{*}{$\underline{0.640} \small{\pm} 0.093$} &  \multirow{1}{*}{$0.628 \small{\pm} 0.073$} &  \multirow{1}{*}{$\underline{0.675} \small{\pm} 0.175$} &  \multirow{1}{*}{$\underline{0.605} \small{\pm} 0.068$} &  \multirow{1}{*}{$0.598 \small{\pm} 0.081$} &    \multirow{1}{*}{$\mathbf{0.629}$}\\

& $\mathbf{\tilde{A}}$~\cite{xiong2021nystrom} &  $0.495 \small{\pm} 0.177$ &  $0.591 \small{\pm} 0.068$ &  $0.600 \small{\pm} 0.190$ &   $0.508 \small{\pm} 0.066$ &   $\underline{0.605} \small{\pm} 0.075$ &    $0.560$ \\
\bottomrule
\end{tabular}
}
\end{table*}

\begin{figure*}[t]
    \centering
    \includegraphics[width=0.99\textwidth]{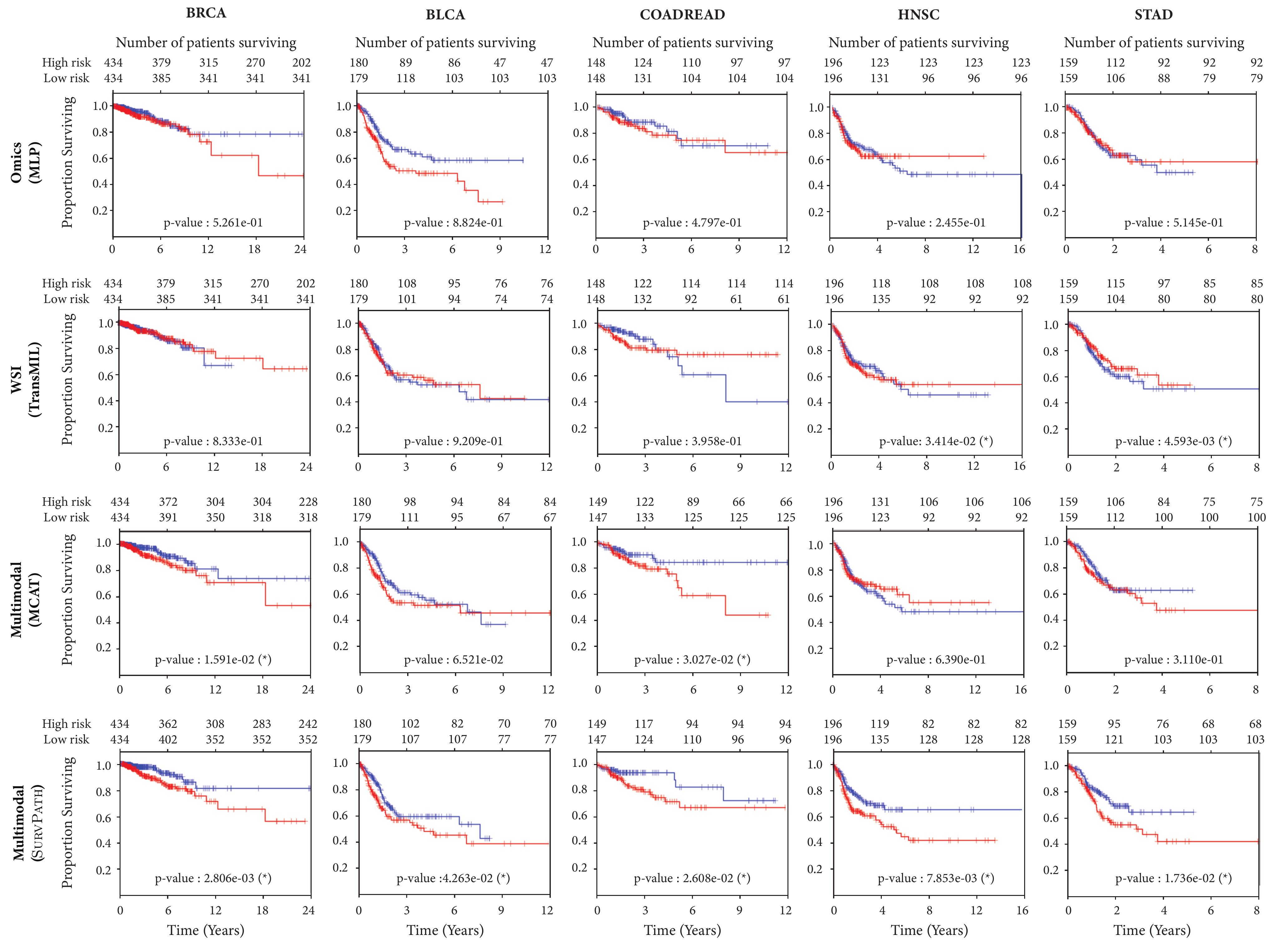}
    \caption{Kaplan Meier curves of $\ours$, compared against histology, transcriptomics, and multimodal baselines. High (red) and low-risk (blue) groups are identified by using the median predicted risk as cut-off. Logrank test was used to determine statistical significance ($\alpha=0.05$).}
    \label{fig:km_curves}
\end{figure*}

\textbf{Kaplan Meier analysis:} Fig.~\ref{fig:km_curves} shows Kaplan-Meier survival curves of predicted high-risk and low-risk groups at 20$\times$. All patients with a risk higher than the median of the entire cohort are assigned as high risk (red), and patients with a risk lower than the median are assigned low risk (blue). For all five diseases, $\ours$ highlights statistically better discrimination of the two risk groups compared to the best histology baseline (TransMIL), transcriptomics baseline (MLP), and multimodal baseline (MCAT). We believe that $\ours$ can better discriminate between risk groups because a simplified early fusion mechanism allows it to find better correlations between transcriptomics and histology with respect to the patient's risk.

\textbf{Comparisons with clinical covariates:} Clinically, prognostication can be based on patient information such as age, and cancer progression assessment, such as cancer grade. We use a Cox proportional hazards model to predict survival from clinical covariates (Age, Sex, Grade) individually and in combination. We find the $\ours$ outperforms survival prediction from all clinical covariates (Table \ref{tab:covariates}). 

\textbf{Modality attributions:} By summing Integrated Gradient (IG) values pre-co-attention per modality, we can derive modality attribution scores (Table~\ref{tab:mod_attr}). We find that histology contributed 77.2\% across cohorts, highlighting the need for multi-modality in prognostication.

\begin{table*}[t]
\caption{Survival prediction results of $\ours$ compared with clinical covariates.  Best performance in \textbf{bold}, second best \underline{underlined}. Predictions using clinical covariates are done using Cox proportional hazards model on the same 5-fold cross-validation splits as $\ours$. } 
\label{tab:covariates}
\centering
\resizebox{1.95\columnwidth}{!}{%
\begin{tabular}{l|cccccc}
\toprule
Model/Study &               BRCA $(\uparrow)$ &               BLCA $(\uparrow)$ &           COADREAD $(\uparrow)$ &               HNSC $(\uparrow)$ &               STAD $(\uparrow)$ &  Overall $(\uparrow)$ \\

\midrule
Age
&  $0.496$$ \small{\pm} 0.086$ &  $\underline{0.578} \small{\pm} 0.056$ &  $0.357 \small{\pm} 0.161$ &  $0.517 \small{\pm} 0.073$ &  $0.499 \small{\pm} 0.055$ & $0.489$\\

Sex                                        
&  $0.490 \small{\pm} 0.011$ &  $0.489 \small{\pm} 0.028$ &  $0.542 \small{\pm} 0.070$ &  $0.486 \small{\pm} 0.035$ &  $0.529 \small{\pm} 0.069$ & $0.507$\\

Grade                                       
&  $\underline{0.597} \small{\pm} 0.078$ &  $0.515 \small{\pm} 0.018$ &  N/A &  $\underline{0.547} \small{\pm} 0.035$ &  $0.552 \small{\pm} 0.055$ &    $0.553$\\

Age + Sex + Grade
&  $0.563 \small{\pm} 0.055$ &    $0.570 \small{\pm} 0.033$ &  $\underline{0.655} \small{\pm} 0.119$ &  $0.512 \small{\pm} 0.093$ &  $\underline{0.592} \small{\pm} 0.044$ &  $\underline{0.578}$\\

$\ours$ (Ours) & $\mathbf{0.655} \small{\pm} 0.089$ &  $\mathbf{0.625} \small{\pm} \mathbf{0.056}$ &  $\mathbf{0.673} \small{\pm} 0.170$ &  $\mathbf{0.600} \small{\pm} 0.061$ &  $\textbf{0.592} \small{\pm} 0.047$ &  $\mathbf{0.629}$ \\

\bottomrule
\end{tabular}
}
\end{table*}


\begin{table*}[t]
\caption{Modality attribution percentages. The sum of Integrated Gradients attribution over all modality-specific tokens before co-attention. Scores reported on validation fold with the highest c-index.} 
\label{tab:mod_attr}
\centering
\resizebox{1.5\columnwidth}{!}{%
\begin{tabular}{l|ccccc}
\toprule
Modality/Study &               BRCA &               BLCA &           COADREAD &               HNSC &               STAD   \\

\midrule
WSI
&  $0.621$$ \small{\pm} 0.251$ &  $0.511 \small{\pm} 0.222$ &  $0.971 \small{\pm} 0.067$ &  $0.921 \small{\pm} 0.060$ &  $0.849 \small{\pm} 0.221$\\

Omics                                        
&  $0.379 \small{\pm} 0.251$ &  $0.489 \small{\pm} 0.222$ &  $0.029 \small{\pm} 0.067$ &  $0.079 \small{\pm} 0.060$ &  $0.151 \small{\pm} 0.221$ \\

\bottomrule
\end{tabular}
}
\end{table*}

\clearpage
\clearpage
{
\small
}

\end{document}